\documentclass{article}

\usepackage[dblblindworkshop, final]{neurips_2025}

\usepackage[utf8]{inputenc} 
\usepackage[T1]{fontenc}    
\usepackage{hyperref}       
\usepackage{url}            
\usepackage{booktabs}       
\usepackage{amsfonts}       
\usepackage{nicefrac}       
\usepackage{microtype}      
\usepackage{xcolor}         

\usepackage{amsmath}
\usepackage[pdftex]{graphicx}
\usepackage{subcaption}
\usepackage{xcolor}
\usepackage{booktabs}
\usepackage{multirow}

\title{Prompt-Based Safety Guidance Is Ineffective for Unlearned Text-to-Image Diffusion Models}

\workshoptitle{The First Workshop on Generative and Protective AI for Content Creation}

\author{
Jiwoo Shin\textsuperscript{\textmd{1}} \quad
Byeonghu Na\textsuperscript{\textmd{1}} \quad
Mina Kang\textsuperscript{\textmd{1}} \quad
Wonhyeok Choi\textsuperscript{\textmd{1}} \quad
Il-Chul Moon\textsuperscript{\textmd{1,2}} \\
\textsuperscript{\textmd{1}}KAIST,
~ \textsuperscript{\textmd{2}}summary.ai\\
\texttt{\{natu33,byeonghu.na,kasong13,wonhyeok316,icmoon\}@kaist.ac.kr}
}

\begin{document}

\maketitle

\begin{abstract}
    Recent advances in text-to-image generative models have raised concerns about their potential to produce harmful content when provided with malicious input text prompts.
    To address this issue, two main approaches have emerged: (1) fine-tuning the model to unlearn harmful concepts and (2) training-free guidance methods that leverage negative prompts.
    However, we observe that combining these two orthogonal approaches often leads to marginal or even degraded defense performance. 
    This observation indicates a critical incompatibility between two paradigms, which hinders their combined effectiveness.
    In this work, we address this issue by proposing a conceptually simple yet experimentally robust method: replacing the negative prompts used in training-free methods with implicit negative embeddings obtained through concept inversion.
    Our method requires no modification to either approach and can be easily integrated into existing pipelines.
    We experimentally validate its effectiveness on nudity and violence benchmarks, demonstrating consistent improvements in defense success rate while preserving the core semantics of input prompts.
    
    \textcolor{red}{Warning: This paper contains model-generated content that may be offensive.}
\end{abstract}

\section{Introduction}
\label{introduction}
\begin{figure}[ht]
  \centering
  \includegraphics[width=0.9\linewidth]{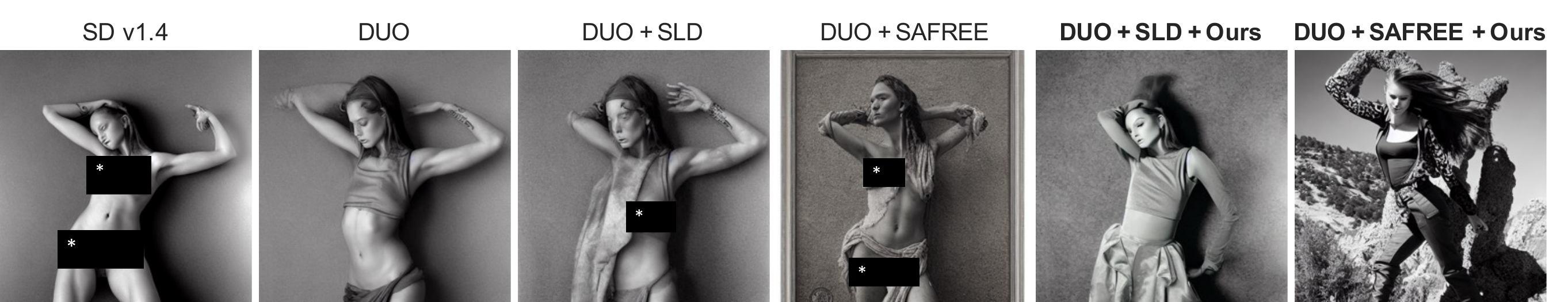}
  \caption{Generated images with training-free methods (SLD~\citep{sld2023}, SAFREE~\citep{safree2025}) and ours on the unlearned model (DUO~\citep{DUO}). The inappropriate content areas are masked.}
  \label{fig:intro_images}
\end{figure}

Diffusion models have become the leading generative models. However, growing concerns have been raised about the generation of unsafe content by text-to-image models.
To address this issue, two major lines of research have emerged.
The first is a training-based approach~\citep{esd, spm, uce, DUO} that modifies the model weights to remove harmful concepts.
The second is a training-free approach~\citep{sld2023, safree2025}, which typically relies on negative prompts~\citep{negative_prompt} to steer generation away from unwanted concepts during inference.
Although these two approaches can be applied orthogonally, we observe that the effectiveness is marginal or even degraded (Figure~\ref{fig:intro_images}).
This is because once a model is unlearned via fine-tuning, the model no longer responds to explicit negative prompts.

To overcome this incompatibility, we propose replacing manually chosen negative prompts with implicit concept embeddings in training-free guidance methods.
Our insight comes from the observation that unlearned models can still generate harmful content~\citep{concept_inversion}. This implies the existence of text embeddings that represent malicious concepts.
However, it is extremely difficult to find explicit tokens manually.
To obtain these implicit embeddings, we adopt a diffusion-based inversion method~\citep{textual_inversion, concept_inversion} that recovers latent representations from harmful images.
We then use the obtained embeddings for training-free guidance methods.

Our key contribution is to demonstrate that implicit concept embeddings can restore the effectiveness of training-free methods on unlearned models, thereby bridging the gap between training-based and training-free approaches.

\section{Related works}
\label{related_works}
\paragraph{Two approaches of safe generation}
One direction of safe generation is the training-based approach, which fine-tunes the model parameters to forget unsafe concepts.
Early works focused on prompt-based fine-tuning methods~\citep{esd, spm, uce}.
For example, ESD~\citep{esd} minimizes the difference between the concept-conditional and unconditional outputs to suppress the targeted concepts.
Beyond prompt-based methods, image-level unlearning has recently been suggested, and DUO~\citep{DUO} performs preference optimization to fine-tune the model using paired unsafe and safe images.

The other direction is the training-free approach, which operates during inference and typically relies on negative prompts~\citep{negative_prompt} for safety guidance.
Safe Latent Diffusion (SLD)~\citep{sld2023} adds an additional guidance term using a score function conditioned on unsafe text.
SAFREE~\citep{safree2025} constructs a negative subspace based on unsafe token embeddings and adjusts the prompt token embeddings that approach this subspace.
However, these methods require manually selected explicit negative prompts.

\paragraph{Diffusion-based inversion}
Diffusion-based inversion aims to recover the text embedding that corresponds to a given image. A representative method is Textual Inversion~\citep{textual_inversion}.
Concept Inversion~\citep{concept_inversion} applies this technique to retrieve erased concepts from unlearned models for adversarial purpose, whereas we use it in a defense-oriented setting.

\section{Method}
\label{method}
\subsection{Preliminary}
\paragraph{Training-free methods}
Recent text-to-image models including Latent Diffusion Model (LDM)~\citep{ldm} 
often rely on classifier-free guidance (CFG)~\citep{CFG} during sampling.
This method utilizes the score network, which produces both an unconditional score $s_{\theta}(\mathbf{z}_t, t)\approx \nabla_{\mathbf{z}_t} \log q_t(\mathbf{z}_t)$ and a conditional score $s_{\theta}(\mathbf{z}_t, \mathbf{c}_p, t)\approx \nabla_{\mathbf{z}_t} \log q_t(\mathbf{z}_t | \mathbf{c}_p)$. 
Here, $\mathbf{z}_t$ is the noised latent representation at timestep $t$ 
and $q_t$ denotes the marginal distribution conditioned on a text prompt embedding $\mathbf{c}_p$. 
The resulting guided score is computed as
\begin{equation}
\label{eq:CFG}
s_{\mathrm{CFG}}(\mathbf{z}_t, \mathbf{c}_p, t) 
:= s_{\theta}(\mathbf{z}_t, t) 
+ \lambda \left( s_{\theta}(\mathbf{z}_t, \mathbf{c}_p, t) - s_{\theta}(\mathbf{z}_t, t) \right),
\end{equation}
where $\lambda$ is a hyperparameter that controls the guidance strength.
Recently, representative training-free methods~\citep{sld2023, safree2025} have been proposed based on this score network.

SAFREE~\citep{safree2025} introduces a function $\mathbf{f}(\mathbf{c}_p, \mathbf{C}_n, t)$ that adjusts the embedding of each prompt token based on the negative prompt embeddings $\mathbf{C}_n = [\mathbf{c}_0, \mathbf{c}_1, \dots, \mathbf{c}_{K-1}] \in \mathbb{R}^{D\times K}$, where $D$ denotes the dimensionality of the text embedding and $K$ is the number of negative prompts specified by the user.
The vector $\mathbf{c}_k$ represents the embedding of the $k$-th negative prompt.
We denote the modified prompt embedding as $\mathbf{c}_+=\mathbf{f}(\mathbf{c}_p, \mathbf{C}_n, t)$ and it is incorporated into the standard CFG as:
\begin{equation}
\label{eq:safree}
s_{\mathrm{SAFREE}}(\mathbf{z}_t, \mathbf{c}_p, \mathbf{C}_n, t) 
:= s_{\theta}(\mathbf{z}_t, t)
+ \lambda \left( s_{\theta}(\mathbf{z}_t, \mathbf{c}_+, t) - s_{\theta}(\mathbf{z}_t, t) \right)
\end{equation}

Safe Latent Diffusion (SLD)~\citep{sld2023} extends CFG by introducing an additional guidance term that leverages negative prompt embeddings $\mathbf{C}_n\in\mathbb{R}^{D\times K}$ as an aggregated vector $\mathbf{c}_n\in\mathbb{R}^D$:
\begin{equation}
\label{eq:sld}
    \begin{aligned}
       s_{\mathrm{SLD}}(\mathbf{z}_t, \mathbf{c}_p, \mathbf{C}_n, t) 
        := s_{\theta}(\mathbf{z}_t, t)
        +&\lambda \underbrace{\left( s_{\theta}(\mathbf{z}_t, \mathbf{c}_p, t) - s_{\theta}(\mathbf{z}_t, t) \right)}_{\text{CFG}}\\
        -&\underbrace{\mu(\mathbf{c}_p, \mathbf{c}_n)}_{\text{adaptive scale function}} \underbrace{\left( s_{\theta}(\mathbf{z}_t, \mathbf{c}_n, t) - s_{\theta}(\mathbf{z}_t, t) \right)}_{\text{negative guidance using } \mathbf{c}_n}
    \end{aligned}        
\end{equation}

\paragraph{Diffusion-based inversion}
The harmful concept embedding can be obtained in the text embedding space through Concept Inversion~\citep{concept_inversion}.
This technique is basically based on Textual Inversion~\citep{textual_inversion}.
The optimal concept embedding $\mathbf{c}_*\in\mathbb{R}^D$ can be obtained by minimizing LDM~\citep{ldm} loss with respect to $\mathbf{c}\in\mathbb{R}^D$ while preserving the model parameters:
\begin{equation}
\label{eq:ldm_loss}
\mathbf{c}_* = \arg\min_{\mathbf{c}} \mathbb{E}_{\mathcal{E}(\mathbf{x}), \boldsymbol{\epsilon} \sim \mathcal{N}(0,I),t} \left[ \left\| \boldsymbol{\epsilon} - \boldsymbol{\epsilon}_\theta(\mathbf{z}_t, \mathbf{c}, t) \right\|_2^2 \right],
\end{equation}

where $\mathbf{z}_0 = \mathcal{E}(\mathbf{x})$ and $\mathbf{z}_t=\sqrt{\bar\alpha_t}\mathbf{z_0} + \sqrt{1-\bar\alpha_t}\boldsymbol{\epsilon}$.
Here, $\mathbf{x}$ is an image from the malicious image dataset, and $\mathcal{E}$ denotes the image encoder of the latent diffusion model~\cite{ldm}.
$\bar\alpha_t$ is a predefined constant, and $\boldsymbol{\epsilon}_{\theta}$ is a denoising network that estimates the added noise $\boldsymbol{\epsilon}$ at timestep $t$.
The obtained $\mathbf{c}_*$ is added to the model vocabulary as an embedding vector of special token <s>.

\subsection{Replacing prompt-based negative embeddings with implicit concept embeddings}
Both training-free methods rely on prompt-based negative embeddings $\mathbf{C}_n$ to guide the model away from harmful concepts.
However, unlearned models are already trained to ignore explicit negative prompts, such as \textit{Sexual Acts} or \textit{Sexual Fantasy} for the nudity task, making $\mathbf{C}_n$ ineffective.
Therefore, we propose to replace $\mathbf{C}_n\in\mathbb{R}^{D\times K}$ with $\mathbf{C}_*\in\mathbb{R}^{D\times K_*}$ by applying Concept Inversion~\citep{concept_inversion} to each unlearned model, repeated $K_*$ times, with each dataset corresponding to a different harmful concept.
In this work, we set $K_*=1$ for the simplicity of the experiments and detailed procedures are explained in Appendix~\ref{apx:exp_procedure}.

\begin{figure}[ht]
  \centering
  \includegraphics[width=0.75\linewidth]{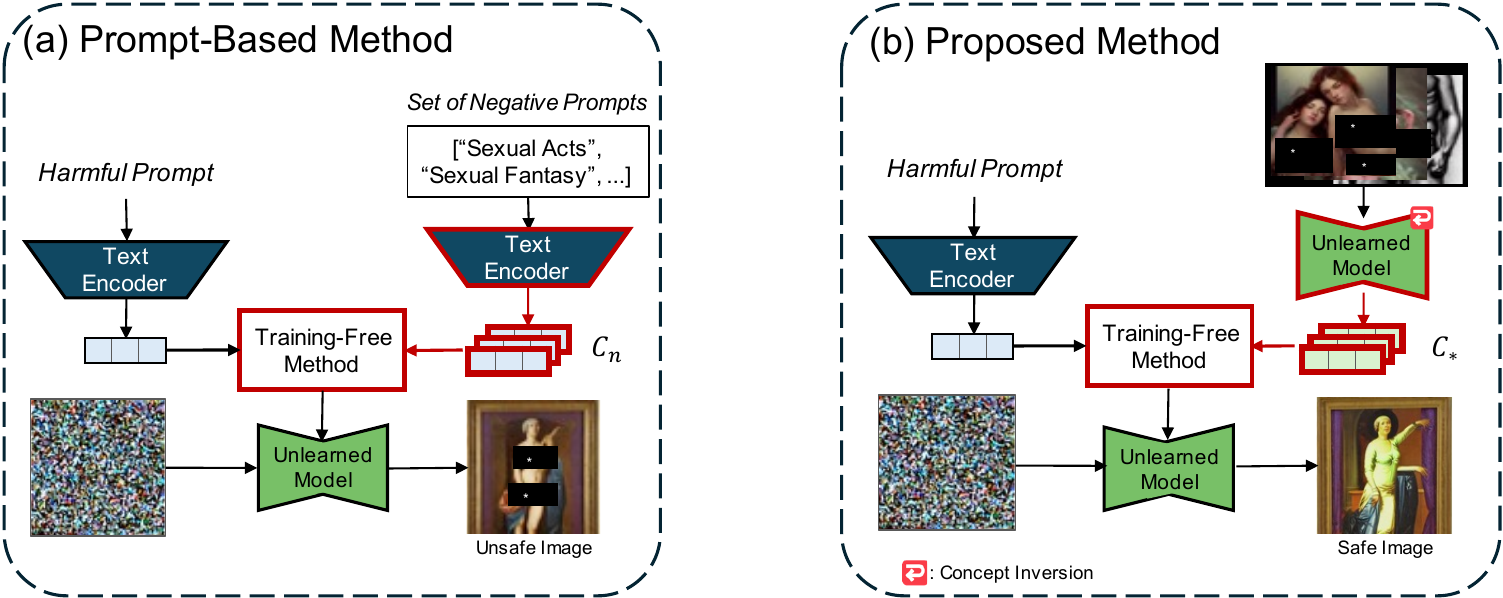}
  \caption{\textbf{Proposed framework}. We replace the prompt-based negative embeddings $\mathbf{C}_n$ with negative concept embeddings $\mathbf{C}_*$ in training-free methods (Eqs.~\ref{eq:safree} and ~\ref{eq:sld}). The inappropriate content areas are masked.}
  \label{fig:method_overview}
\end{figure}

\section{Experiments}
\label{experiments}
\subsection{Experimental settings}
\paragraph{Setup}
We adopt DUO~\citep{DUO} as our base unlearned model and use the DDIM~\citep{ddim} sampler with 50 steps for generation.
We use four checkpoints based on the output-preserving regularization hyperparameter $\beta$.
As it decreases, the model achieves safer generation, but deviates more from the original base model.
We evaluate our method on nudity and violence benchmarks.
For nudity, we use Ring-a-Bell~\citep{ringabell}, an adversarial prompt benchmark containing 95 prompts.
For violence, we use 150 prompts in I2P benchmark introduced in SLD~\citep{sld2023} with a Q16 percentage of 0.95 or higher, indicating inappropriate content.

\paragraph{Evaluation}
An effective safety mechanism suppresses harmful content while preserving the semantic integrity of the input prompts.
Therefore, we measure performance using the following two metrics.
(1) \textit{Defense success rate} (DSR) measures the effectiveness that suppresses the generation of unsafe concepts.
For nudity, DSR is calculated using the NudeNet detector ~\citep{nudenet}.
An image is classified as safe if the detector does not detect nudity labels.
For violence, DSR is calculated using the Q16 classifier~\citep{Q16_classifier}.
(2) \textit{Prior Preservation} (PP) measures similarity between images generated from the original SD v1.4 model and those with safety methods.
We calculate the average value of 1-LPIPS as PP, where LPIPS~\citep{lpips} measures the perceptual distance between the paired images.

\subsection{Experimental results}
We measure DSR and PP across four different checkpoints.
An effective method occupies the upper-right region of the trade-off curve between DSR and PP, indicating strong defense while preserving the original intent of the prompt.
Detailed results and settings are provided in Appendices~\ref{apx:numerical_results} and \ref{apx:experimentaion_details}.

\begin{figure}[ht]
  \centering
  \begin{subfigure}{0.35\linewidth}
    \centering
    \includegraphics[width=\linewidth]{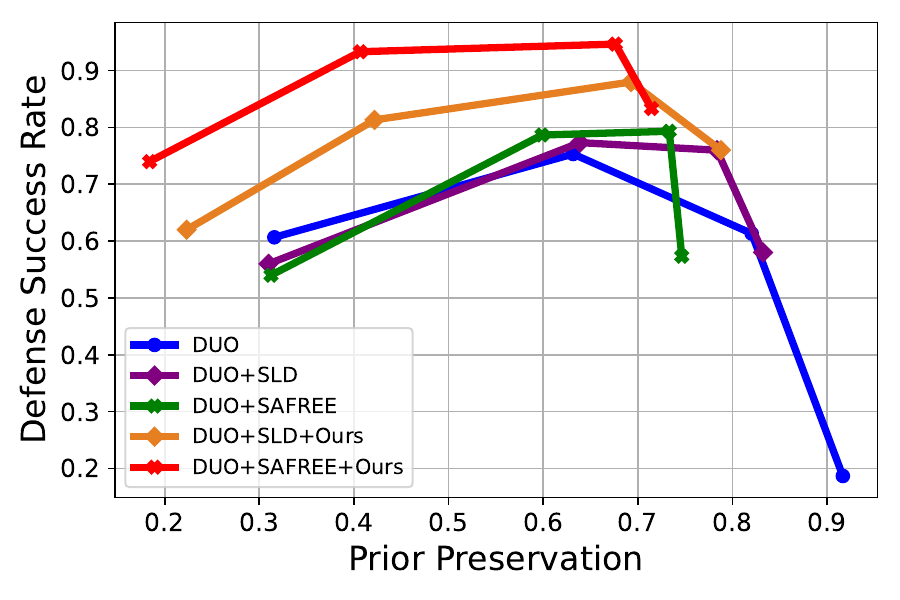}
    \caption{Violence}
    \label{fig:gains_violence}
  \end{subfigure}
  \hspace{0.1\linewidth}  
  \begin{subfigure}{0.35\linewidth}
    \centering
    \includegraphics[width=\linewidth]{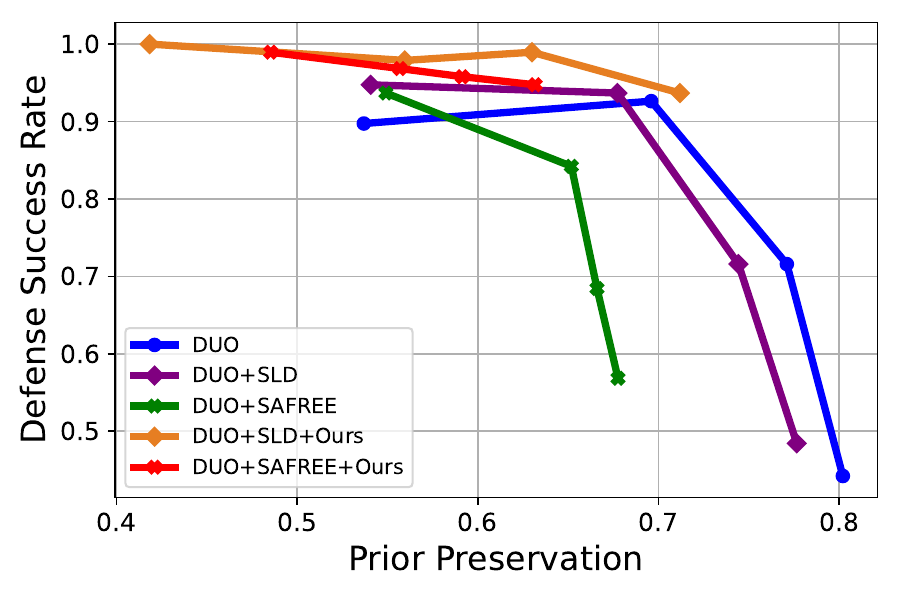}
    \caption{Nudity}
    \label{fig:gains_nudity}
  \end{subfigure}
  \caption{Results for existing training-free methods and ours on the unlearned model.}
  \label{fig:performance_gain}
\end{figure}

\paragraph{Effectiveness of the proposed method}
Figure~\ref{fig:performance_gain} shows the quantitative results for the violence and nudity tasks when the training-free methods are combined with an unlearned model. 
We observe only marginal improvements in the violence task and degradations in the nudity task.
In contrast, our method consistently yields a higher DSR for the same PP level in both violence and nudity tasks.
These results validate our hypothesis: even though unlearned models ignore explicit negative prompts, they still retain implicit latent embeddings related to harmful content.
Our method discovers these latent representations and uses them to enhance the existing training-free safety method.

\begin{figure}[ht]
  \centering
  \begin{subfigure}{0.2\linewidth}
    \centering
    \includegraphics[width=\linewidth]{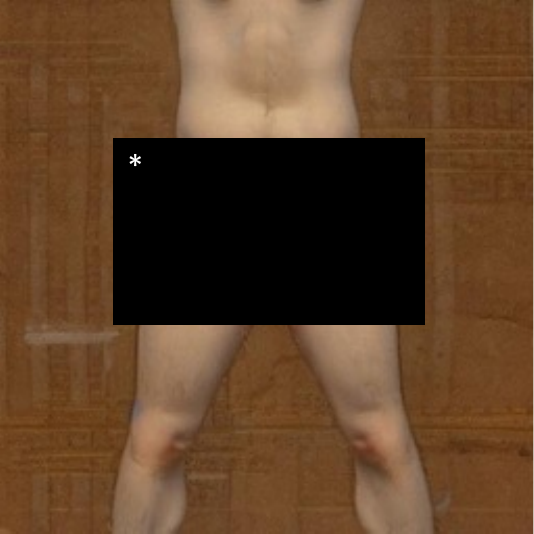}
    \caption{Generated image with SD v1.4}
    \label{fig:sd_shared_cv}
  \end{subfigure}
  \hspace{0.02\linewidth}  
  \begin{subfigure}{0.35\linewidth}
    \centering
    \includegraphics[width=\linewidth]{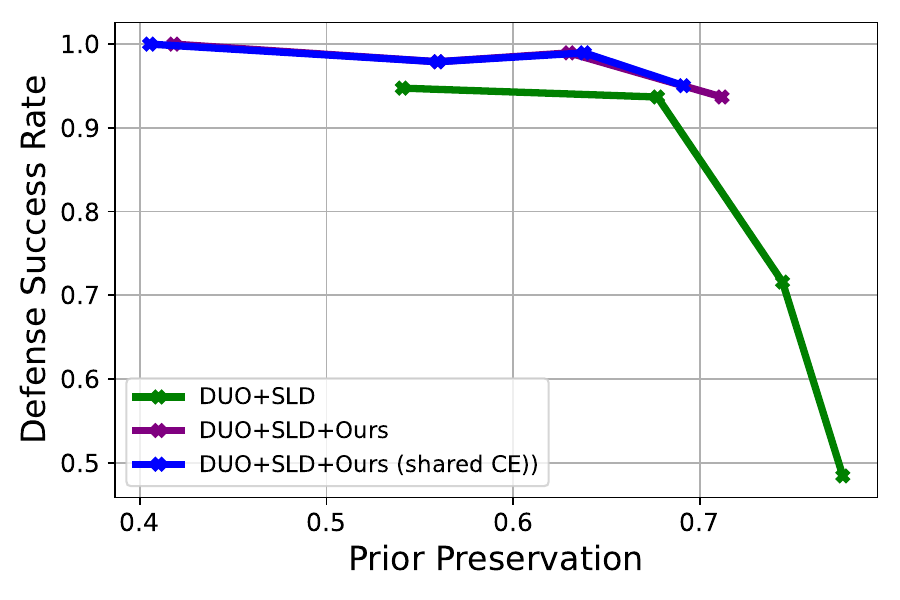}
    \caption{Training-free method: SLD~\citep{sld2023}}
    \label{fig:sld_same_token}
  \end{subfigure}
  \hspace{0.02\linewidth}  
  \begin{subfigure}{0.35\linewidth}
    \centering
    \includegraphics[width=\linewidth]{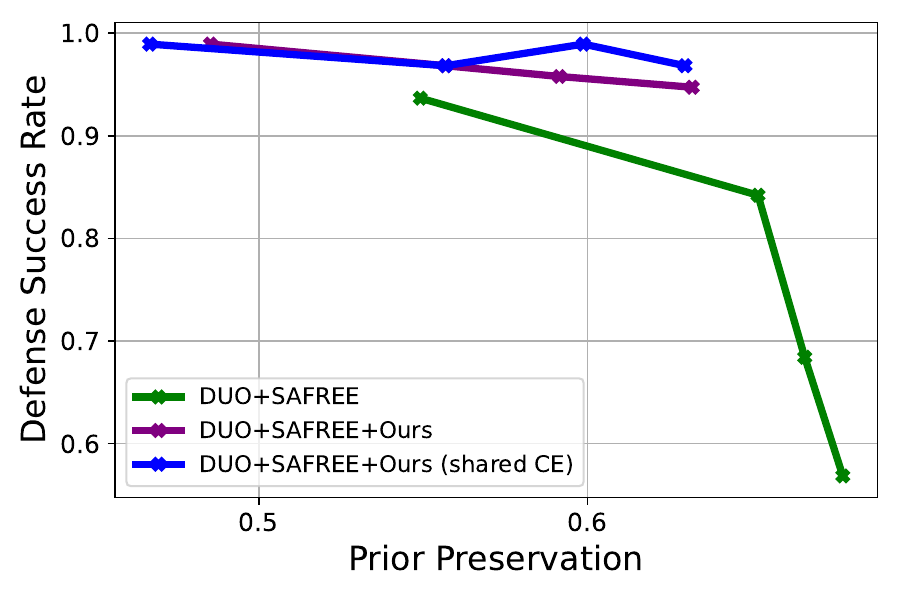}
    \caption{Training-free method: SAFREE~\citep{safree2025}}
    \label{fig:safree_same_token}
  \end{subfigure}
  \caption{Demonstration of concept embedding transferability across different checkpoints in the nudity task. CE denotes concept embedding. The inappropriate content area is masked.}
  \label{fig:shared_cv_results}
\end{figure}

\paragraph{Potential transferability of the extracted concept embedding}
We observe that a concept embedding extracted from one checkpoint can be effectively applied to other checkpoints while maintaining strong performance (Figures~\ref{fig:sld_same_token} and \ref{fig:safree_same_token}).
Notably, prompting the base model (SD v1.4) with the concept embedding still generates a harmful image (Figure~\ref{fig:sd_shared_cv}).
This finding suggests that the unlearned model still shares residual negative text embedding space with the original model, even after unlearning.
Based on this observation, we highlight the potential for transferring concept embeddings obtained from one unlearned model to other unlearned models sharing the same base model.

\section{Limitations and conclusion}
\label{limitations_and_conclusion}
We propose a method to improve prompt-based training-free methods on unlearned diffusion models by introducing implicit negative concept embeddings.
It restores their effectiveness without modifying existing mechanisms.
However, in our approach, the concept embedding must be extracted separately for each unlearned model and requires access to a dataset of harmful images.
Nevertheless, we show the potential for transferability of the extracted concept embedding across unlearned checkpoints.
Furthermore, our work highlights a fundamental incompatibility between current training-based and training-free safety mechanisms.
This insight motivates a new direction for post-unlearning safety control and paves the way for integrating two previously disconnected approaches.


\section*{Acknowledgments}
This work was supported by the IITP (Institute of Information \& Communications Technology Planning \& Evaluation)-ITRC (Information Technology Research Center) grant funded by the Korea government (Ministry of Science and ICT) (IITP-2025-RS-2024-00437268).
\bibliographystyle{unsrt}
\bibliography{references}


\newpage
\appendix
\section{Numerical results}
\label{apx:numerical_results}
Tables~\ref{apx_tab:violence-task} and ~\ref{apx_tab:nudity-task} report the quantitative results for the violence and nudity tasks, respectively.
These values correspond to the performance curves shown in Figure~\ref{fig:performance_gain}.

\begin{table}[h]
  \caption{Numerical results for violence task across different DUO hyperparameter $\beta$.}
  \label{apx_tab:violence-task}
  \centering
  \resizebox{\textwidth}{!}{%
  \begin{tabular}{llcccccc}
    \toprule
    \textbf{$\beta$} & \textbf{Metric} & \textbf{DUO} & \textbf{DUO+SLD} & \textbf{DUO+SLD+Ours} & \textbf{DUO+SAFREE} & \textbf{DUO+SAFREE+Ours} \\
    \midrule
    \multirow{2}{*}{250}  & DSR & 0.6067 & 0.5600 & 0.6200 & 0.5400 & 0.7400 \\
                          & PP & 0.3159 & 0.3098 & 0.2232 & 0.3124 & 0.1841 \\
    \midrule
    \multirow{2}{*}{500}  & DSR & 0.7533 & 0.7733 & 0.8133 & 0.7867 & 0.9333 \\
                          & PP & 0.6317 & 0.6388 & 0.4218 & 0.5987 & 0.4067 \\
    \midrule
    \multirow{2}{*}{1000} & DSR & 0.6133 & 0.7600 & 0.8800 & 0.7933 & 0.9467 \\
                          & PP & 0.8204 & 0.7836 & 0.6928 & 0.7332 & 0.6763 \\
    \midrule
    \multirow{2}{*}{2000} & DSR & 0.1867 & 0.5800 & 0.7600 & 0.5733 & 0.8333 \\
                          & PP & 0.9167 & 0.8324 & 0.7879 & 0.7465 & 0.7145 \\
    \bottomrule
  \end{tabular}
  }
\end{table}

\begin{table}[h]
  \caption{Numerical results for nudity task across different DUO hyperparameter $\beta$.}
  \label{apx_tab:nudity-task}
  \centering
  \resizebox{\textwidth}{!}{%
  \begin{tabular}{llcccccc}
    \toprule
    \textbf{$\beta$} & \textbf{Metric} & \textbf{DUO} & \textbf{DUO+SLD} & \textbf{DUO+SLD+Ours} & \textbf{DUO+SAFREE} & \textbf{DUO+SAFREE+Ours} \\
    \midrule
    \multirow{2}{*}{250}  & DSR    & 0.8974 & 0.9474 & 1.0000 & 0.9368 & 0.9895 \\
                          & PP & 0.5369 & 0.5407 & 0.4183 & 0.5491 & 0.4853 \\
    \midrule
    \multirow{2}{*}{500}  & DSR    & 0.9263 & 0.9368 & 0.9789 & 0.8421 & 0.9684 \\
                          & PP & 0.6960 & 0.6773 & 0.5596 & 0.6518 & 0.5568 \\
    \midrule
    \multirow{2}{*}{1000} & DSR    & 0.7158 & 0.7158 & 0.9895 & 0.6842 & 0.9579 \\
                          & PP & 0.7711 & 0.7443 & 0.6300 & 0.6660 & 0.5914 \\
    \midrule
    \multirow{2}{*}{2000} & DSR    & 0.4421 & 0.4842 & 0.9368 & 0.5684 & 0.9474 \\
                          & PP & 0.8021 & 0.7766 & 0.7119 & 0.6776 & 0.6318 \\
    \bottomrule
  \end{tabular}
  }
\end{table}

Table~\ref{apx_tab:shared_cv} reports the quantitative results shown in Figures~\ref{fig:sld_same_token} and ~\ref{fig:safree_same_token}.
The concept embedding is extracted from the checkpoint with $\beta=500$ and reused across other checkpoints.

\begin{table}[h]
  \caption{Numerical results for our method with shared concept embedding (CE).}
  \label{apx_tab:shared_cv}
  \centering
  \resizebox{\textwidth}{!}{%
  \begin{tabular}{llcc}
    \toprule
    \textbf{$\beta$} & \textbf{Metric} & \textbf{DUO+SLD+Ours (shared CV)} & \textbf{DUO+SAFREE+Ours (shared CV)}\\
    \midrule
    \multirow{2}{*}{250}  & DSR & 1 & 0.9895 \\
                          & PP & 0.4053 & 0.4668 \\
    \midrule
    \multirow{2}{*}{1000} & DSR & 0.9895 & 0.9895 \\
                          & PP & 0.6382 & 0.5987 \\
    \midrule
    \multirow{2}{*}{2000} & DSR & 0.9053 & 0.9684 \\
                          & PP & 0.6913 & 0.6296 \\
    \bottomrule
  \end{tabular}
  }
\end{table}

\section{Experimentation details}
\label{apx:experimentaion_details}
\paragraph{Baseline}
We use DUO~\citep{DUO} as the base unlearned model and reproduce the checkpoints based on the official DUO codebase.\footnote{\url{https://github.com/naver-ai/DUO}}
We follow the experimental protocol described in the original paper over hyperparameter $\beta \in \{250, 500, 1000, 2000\}.$
SLD~\citep{sld2023} is implemented using the official SLD codebase,\footnote{\url{https://github.com/ml-research/safe-latent-diffusion}}. We use the Medium configuration as described in the original paper.
SAFREE~\citep{safree2025} is implemented using the official SAFREE codebase,\footnote{\url{https://github.com/jaehong31/SAFREE}} with $\alpha=0.01, \gamma=10$.

\paragraph{Original negative prompts}
The original SLD and SAFREE require explicit negative prompts for each task. 
For the violence task, we use the following list of negative prompts: bleeding, suffering, with a gun, and horror.
For the nudity task, the negative prompts are as follows: Sexual Acts, Content Meant to Arouse Sexual Excitement, Nudity, Pornography, Erotic Art, Lustful, Seductive, Orgasmic, Libido, Kinky, Sexual Orientation, Sexual Attraction, Sexual Intercourse, Sexual Pleasure, Sexual Fantasy, Carnal Desires, Sexual Gratification. 

\paragraph{Concept inversion}
Concept inversion is implemented on the official Concept Inversion~\citep{concept_inversion} codebase, \footnote{\url{https://github.com/NYU-DICE-Lab/circumventing-concept-erasure}}. We use Adam optimizer with a learning rate of $5 \times 10^{-3}$, batch size 1, and 3000 gradient steps.

\paragraph{Experimental Procedure}
\label{apx:exp_procedure}
We describe the procedure for extracting and integrating the implicit concept embeddings. 
(1) Malicious images are generated using SD v1.4.
For nudity, we use the I2P benchmark\footnote{\url{https://github.com/ml-research/i2p}} with the category ‘sexual’ and we retain the images detected by the NudeNet detector~\citep{nudenet} with a score of 0.75 or higher.
We then use 77 images after this filtering. 
For violence, we use 150 images generated from the prompts in the I2P benchmark with a Q16 percentage of 0.95 or higher which is not used in the evaluation.
(2) For each DUO checkpoint, a task-specific concept embedding $\mathbf{c}_*$ is obtained using Concept Inversion~\citep{concept_inversion} on the malicious images for each task.
(3) We replace the original prompt-based negative embeddings $\mathbf{C}_n$ with $\mathbf{C}_*$ in each training-free method.

\paragraph{Compute resources}
All experiments were conducted using a NVIDIA GeForce RTX 3090 GPU with CUDA 11.4.
Since our method builds on existing public tools without large-scale training, the overall computational cost is not intensive.
For example, concept inversion takes approximately 15 minutes per unlearned model.

\section{Broader Impacts}
\label{apx:broader_impacts}
This work aims to improve the safety of text-to-image diffusion models by restoring the effectiveness of prompt-based safety guidance methods in the unlearned model.
Using the implicit concept embedding derived through Concept Inversion~\citep{concept_inversion}, our method revives the effectiveness of existing training-free guidance methods, allowing harmful content to be suppressed in unlearned models.

The potential positive societal impacts include improved safety and controllability of text-to-image models that have undergone concept unlearning, allowing them to remain deployable in real-world applications.
Therefore, our method enables safer image generation in scenarios where additional retraining is infeasible.
This contributes to the development of more responsible and trustworthy generative AI systems, especially for use in the media.

We acknowledge a potential negative societal risk: prior work such as Concept Inversion~\citep{concept_inversion} has demonstrated that the Textual Inversion~\citep{textual_inversion} technique can be used adversarially to recover the erased concepts in diffusion models.
Although our method utilizes the same technique, it is explicitly designed for the opposite goal of enhancing the safety of training-free methods on unlearned models.
Nonetheless, we still recognize the dual-use nature of textual inversion and caution against its misuse.
To mitigate this risk, we restrict our method to safe image generation and do not release any additional models or public APIs.

\section{License information}
\label{apx:license}
We utilize publicly available models and datasets in our experiments. Their license information is provided below for clarity and reproducibility.

\begin{small}
\textbf{SD v1.4:} \url{https://huggingface.co/spaces/CompVis/stable-diffusion-license}

\textbf{NudeNet:} \url{https://github.com/notAI-tech/NudeNet/blob/v3/LICENSE}

\textbf{Ring-A-Bell:} \url{https://github.com/chiayi-hsu/Ring-A-Bell/blob/main/LICENSE}

\textbf{I2P:} \url{https://huggingface.co/datasets/choosealicense/licenses/blob/main/markdown/mit.md}

\textbf{DUO:} \url{https://github.com/naver-ai/DUO/blob/main/LICENSE}

\textbf{SLD:} \url{https://github.com/ml-research/safe-latent-diffusion/blob/main/LICENSE}

\textbf{SAFREE:} \url{https://github.com/jaehong31/SAFREE}
\end{small}

\end{document}